\documentclass[letterpaper, 10 pt, conference]{Attachment/ieeeconf}

\IEEEoverridecommandlockouts                              
                                                          
\usepackage[utf8]{inputenc}
\usepackage{amsmath}
\usepackage{mathtools}
\usepackage{amsfonts}
\usepackage{xcolor}
\usepackage[english]{babel}
\usepackage{stackengine}
\usepackage{amssymb}
\usepackage{mathrsfs}
\usepackage{algorithm, float}
\usepackage{algorithmic}
\usepackage{bbm}
\usepackage{dsfont}
\usepackage{booktabs} 
\usepackage{caption}
\usepackage{subcaption}
\usepackage{lipsum}
\usepackage{graphicx}

\newtheorem{The}{Theorem}
\newtheorem{ass}{Assumption}
\newtheorem{lemma}{Lemma}

\newtheorem{Pro}{Problem}

\def\Var{{\textrm{Var}}\,}

\DeclareMathOperator{\E}{\mathbb{E}}

\title{RRT Guided Model Predictive Path Integral Method}

\author{Chuyuan Tao$^{\dagger}$, Hunmin Kim$^{\ddagger}$, and  Naira Hovakimyan$^{\dagger}$
\thanks{This research is supported by NSF CPS \#1932529, AFOSR \#FA9550-21-1-0411,
  NASA \#80NSSC22M0070 and \#80NSSC20M0229 awards.}
\thanks{$^{\dagger}$Chuyuan Tao and Naira Hovakimyan are with the Department of Mechanical Science and Engineering, University of Illinois at Urbana-Champaign, USA.
{\tt\small  \{chuyuan2, nhovakim\}@illinois.edu}}%
\thanks{$^{\ddagger}$Hunmin Kim is with the Department of Electrical and Computer Engineering, Mercer University, USA.
{\tt\small  kim$\_$h@mercer.edu }}
}

\begin{document}

\maketitle
\thispagestyle{plain}
\pagestyle{plain}
\begin{abstract}
This work presents an optimal sampling-based method to solve the real-time motion planning problem in static and dynamic environments, exploiting the Rapid-exploring Random Trees (RRT) algorithm and the Model Predictive Path Integral (MPPI) algorithm. The RRT algorithm provides a nominal mean value of the random control distribution in the MPPI algorithm, resulting in satisfactory control performance in static and dynamic environments without a need for fine parameter tuning. We also discuss the importance of choosing the right mean of the MPPI algorithm, which balances exploration and optimality gap, given a fixed sample size. In particular, a sufficiently large mean is required to explore the state space enough, and a sufficiently small mean is required to guarantee that the samples reconstruct the optimal controls. The proposed methodology automates the procedure of choosing the right mean by incorporating the RRT algorithm. The simulations demonstrate that the proposed algorithm can solve the motion planning problem in real-time for static or dynamic environments.
\end{abstract}

\section{Introduction}

Motion planning problems have been widely discussed in recent years in the field of robotics, such as self-driving car navigation, automatic drone, and bipedal robots~\cite{gao2020teach, zhou2021raptor, nguyen2020dynamic, nguyen2018dynamic,choudhury2021efficient}. The main goal of motion planning problems is to find a path for the agents to move from an initial position to a target position in fully-known environments while preventing collisions. However, it still remains challenging to solve the optimal motion planning problems efficiently in dynamic environments and implement the algorithms on the robotic systems in real-time.

For motion planning problems, sampling-based methods have been proven to be effective for complex systems since the methods avoid calculating the derivatives of the dynamic equation and the cost function. In particular, the Probabilistic Roadmap (PRM) algorithm~\cite{kavraki1996probabilistic} is the first sampling-based algorithm that solves the motion planning problem. The algorithm utilizes a local planner to connect the sampling configuration in free space. The Rapid-exploring Random Trees (RRT) algorithm~\cite{lavalle2001randomized, lavalle1998rapidly}, one of the most famous sampling-based algorithms, combines the exploration of the configuration space and the biased sampling around the goal configuration space. Most of the RRT algorithm variants can efficiently solve motion planning problems but can not find an optimal solution. The RRT* algorithm has been developed in~\cite{karaman2011sampling} to find an optimal solution by using incremental rewirings of the graph to provide an asymptotically optimal solution to the motion planning problems. However, compared to the RRT algorithm, the RRT* algorithm and its variants have a relatively longer execution time because the algorithm calculates the neighboring nodes and rewires the graph.

Most RRT and RRT* algorithms can not handle dynamic environments since it requires one to abandon the current result path, and the new path grows from scratch. Dynamic Rapidly-exploring Random Trees (DRRTs) algorithm~\cite{ferguson2006replanning} was developed to address the problem by trimming the original results and exploring to get the target again. In~\cite{zucker2007multipartite}, the authors provide a variant of replanning RRT algorithms combined with the Multipartite RRT (MP-RRT) algorithm. The MP-RRT algorithm biases the sampling distribution towards previous useful states and analytically computes which part of the previous RRT results can be re-utilized. Yet, both algorithms could not guarantee an optimal solution to the motion planning problem since the algorithms are based on non-optimal RRT algorithms. Thus, we provide a different approach to solving optimal real-time motion problems.

One alternative way to efficiently solve optimal motion planning problems with dynamic environments is to use the Model Predictive Integral Control (MPPI) algorithm~\cite{williams2015model, williams2016aggressive}. By sampling the forward trajectories of dynamic systems, the MPPI algorithm avoids calculating the derivatives of the dynamic functions or the cost functions~\cite{williams2017information}. Since the forward trajectories can be sampled efficiently by Graphic Processing Units (GPUs), the algorithm can be applied to diverse robotic systems with finishing the calculation in a fixed time~\cite{williams2018information}. The MPPI algorithm enables real-time implementation by adjusting the fixed computation time, whereas a longer computation time reduces the optimality gap. Since the algorithm solves the motion planning problem iteratively, the algorithm can handle dynamic environments directly. However, the performance of the algorithm is influenced by the hyper-parameters dramatically, especially the mean value of the control input sample distribution. Intuitively, a small mean value may result in conservative exploration, and a large mean value may result in risky behaviors. Especially in dynamic environments, to get better performance, a time-varying mean value is needed. Thus, in this work, we utilize the RRT algorithm to design a better sample mean to guide the MPPI algorithm in exploring the workspace and sampling the trajectories.

The idea of using the RRT and RRT* algorithms to solve motion planning problems in dynamic environments is inspired by~\cite{otte2016rrtx}. In this work, the authors propose the RRT$^X$ algorithm, which combines the replanning ideas provided in the DRRT algorithm and RRT* algorithm to continuously update the path during the exploration when the environment changes. However, the algorithms require large computation power and are hard to implement on the robots in real-time.

The idea of using a nominal or baseline controller to improve the performance of the MPPI algorithm is inspired by~\cite{hatch2021value, yin2022trajectory}.
In~\cite{hatch2021value}, the authors present a method using the entire planning tree from RRT* algorithm to approximate the value functions in the MPPI algorithm. However, due to the learning procedure in the algorithm, the cost of finding an optimal solution to the motion planning problem is computationally expensive. In~\cite{yin2022trajectory}, the authors control the variance of the MPPI algorithm to handle the dynamic environments and provide a faster running time and better collision avoidance in a ground unicycle simulation. However, the algorithm requires the linearized dynamic model, which results in expensive computation for complex or high-dimension systems.

\textbf{Contribution.}
This work presents a real-time sampling-based algorithm to solve the motion planning problem. We use the RRT algorithm to provide a nominal mean value for the random control distribution of the MPPI algorithm. The proposed algorithm advances the RRT algorithm in terms of dynamic environment navigation and optimality, and with respect to the MPPI algorithm, it reduces the need to fine-tune the mean value. We provide simulation results and sample size analysis to discuss the importance of tuning the mean value in the original MPPI algorithm. In particular, the sampling-based algorithms need sufficiently large means to explore the state space and sufficiently small means to guarantee that the samples reconstruct the optimal control. Thus, our proposed method avoids fine-tuning the mean value by using the nominal path provided by the RRT algorithms.
Our algorithm finds the optimal solutions by allowing the MPPI algorithm to explore freely, and it has a better performance in running time by using the RRT algorithm to provide a nominal path to guide the MPPI algorithm. If the MPPI algorithm reaches the area where the nominal path provided by the RRT algorithm has not been explored before, our algorithm uses a real-time Replanning RRT algorithm to provide new nominal paths. Finally, we implement our algorithm on a unicycle robot and solve the motion planning problem in static and dynamic environments.

The rest of the paper is organized as follows. Section \ref{sec:problem} formulates the optimal motion planning problems. Section \ref{sec:algorithms} proposes the RRT-MPPI algorithm to solve the motion planning problems provided in Section \ref{sec:problem}. We show the necessity of an automated design procedure for an RRT-based MPPI algorithm by demonstrating that a large mean is desired for exploration purposes in Section~\ref{sub:MPPI} and by showing that a small mean is desired to reduce the optimality gap in Section~\ref{sec:analysis}, given fixed sample size. Section \ref{sec:simulations} provides simulations for a ground robot navigating through static and dynamic environments. We also compare the running time of the original MPPI algorithm and our RRT-based MPPI algorithm.


\section{Problem Statement}\label{sec:problem}
Let $\mathcal{X}^d$ denote the $d$ dimensional space and $\mathcal{X}_{obs} \subset \mathcal{X}^d$ be the obstacle space. We define the free space $\mathcal{X}_f = \mathcal{X}^d \setminus \mathcal{X}_{obs}$, where the robot can reach. The start and goal states are $x_s$ and $x_g$, respectively.
We assume that the robot has a nonlinear control affine dynamical system:
\begin{equation}\label{eq:dynamics}
    dx_t = (f(x_t)+ g(x_t)u_t)dt + \sigma(x_t)dW_t,
\end{equation}
where $x_t \in \mathbb{R}^n$ is the state, $f: \mathbb{R}^n \rightarrow \mathbb{R}^n$, $g:\mathbb{R}^n \rightarrow \mathbb{R}^{n\times m}$ and $\sigma: \mathbb{R}^n \rightarrow \mathbb{R}^n$ are locally Lipschitz continuous functions, and $dW_t$ is a Wiener process with $\left < dW_k dW_l \right > = \nu_{kl}(x_t,u_t,t)dt$.

\begin{Pro}\label{pro1}
Given $\mathcal{X}, \mathcal{X}_{obs}, x_g$, and $x_t(0) = x_s$, we aim to find the optimal control input $u^*$ that would lead to the shortest path to state $x_g$ in the static or dynamic environments (i.e., time-varying $\mathcal{X}_{obs}$). Specifically, we aim to minimize the cost functions $S(x_t, u_t)$, defined as:
\begin{equation}\label{eq:prob}
    S(x_t,u_t) = \phi(x_{t+T}) +\sum_{j=t}^{t+T-1} q(x_{j}, u_j)
\end{equation}
subject to $x_j \in \mathcal{X}_f$, where $q(x_{j}, u_j)$ is a running cost function,
and $\phi(x_{t+T})$ is the terminal cost function.
\end{Pro}


\section{Approach}\label{sec:algorithms}

To this end, we propose the RRT-guided MPPI algorithm in Section~\ref{sub:RRT-MPPI}, which addresses Problem~\ref{pro1}. By generating mean values using the RRT (presented in Section~\ref{sub:RRT}), the MPPI algorithm (presented in Section~\ref{sub:MPPI}) does not need fine parameter tuning in dynamic environments. Furthermore, the MPPI algorithm provides real-time implementable optimal control inputs.

\begin{algorithm}
\caption{RRT algorithm} \label{alg:RRT}
\begin{algorithmic}\label{RRT_algo}
\STATE{\textbf{Given:} Initial vertices $\mathcal{S} \leftarrow \left \{ s_{s} \right \}$}
\STATE{\textbf{Given:} Initial edges $\varepsilon \leftarrow \varnothing$}
\FOR{$i=1,...,n$}
    \STATE{$s_{sample}\leftarrow$\textit{SampleState}();}
    \STATE{$s_{near}\leftarrow$\textit{NearestNeighbor}$(\mathcal{S}, s_{sample})$;}
    \STATE{$s\leftarrow$\textit{Steer}$(s_{near},s_{sample}, \gamma)$;}
    \IF{\textit{ObstacleFree$(s_{near}, s)$}}
        \STATE{$\mathcal{S} \leftarrow \mathcal{S} \cup \left \{ s \right \}$;}
        \STATE{$\varepsilon \leftarrow \varepsilon \cup \left \{ (s_{near}, s) \right \}$;}
    \ENDIF
    \IF{$d(s,s_g)<\gamma$}
        \STATE{$\mathcal{S} \leftarrow \mathcal{S} \cup \left \{ s_g \right \}$;}
        \STATE{$\varepsilon \leftarrow \varepsilon \cup \left \{ (s, s_g) \right \}$;}
        \STATE{\textbf{return} $p=$ \textit{ExtractPath} $(\mathcal{S}, \varepsilon)$}
    \ENDIF
\ENDFOR
\end{algorithmic}
\end{algorithm}

\subsection{Rapidly-Exploring Random Trees Algorithm}\label{sub:RRT}

We present the RRT algorithm in Algorithm \ref{RRT_algo}. The algorithm uses function \textit{SampleState}() to uniformly sample a new state $s_{sample}$ in the configuration space $\mathcal{X}$. Then, the algorithm finds the nearest vertex $s_{near}$ with function \textit{NearestNeighbor}() and projects the sample state $s_{sample}$ to the ball with radius $\gamma$ with function \textit{Steer}(). If the new edge between the $s_{near}$ and $s$ is free from collision, the projected state $s$ will be added to the vertex set, and the new edge $(s,s_{nearest})$ will be added to the vertex set. If the new vertex $s$ is within a radius $\gamma$ of the goal state $s_{goal}$, then the RRT reaches the target and returns the path $p$. Otherwise, the algorithm adds the new vertex and advances the exploration.

The RRT algorithm focuses on fast iteration while not guaranteeing finding the optimal solution to the motion planning problem. RRT*~\cite{karaman2011sampling} is the first variant of the RRT algorithm that could ensure asymptotic optimality. By allowing the new vertices to "rewire" graph edges within the local neighborhood, the algorithm guarantees asymptotic optimality with the cost of increasing running time. However, the computation time of the RRT* algorithm also increases dramatically, and it is hard to implement the algorithm in real-time. In this work, instead of using the RRT* algorithm, we utilize the MPPI algorithm to find the optimal solution to the motion planning problems.

\subsection{Model Predictive Path Integral Control Algorithm} \label{sub:MPPI}
In this section, we introduce the MPPI algorithm~\cite{williams2017information, williams2018information} to solve the motion planning problems. First, we need to sample $K$ trajectories with time horizon $T$ with random control input $u_{i,j} \sim \mathcal{N}(\mu, \Sigma)$, where $i=1,\cdots,K$ is the sample trajectory index. In each trajectory $\tau_i$, we denote $u_i=[u_{i,t}, \dots, u_{i,t+T-1}]^T$ the actual control input sequence, and $[x_{i,t+1}, \dots, x_{i,t+T}]^T$ the states of the current sample trajectory. The evaluated cost for $i^{th}$ trajectory is given by
\begin{equation}
S(\tau_i)=\phi(x_{i,t+T})+\sum_{j=t}^{t+T-1} q(x_{i,j},u_{i,j}),
\end{equation}
where $q(x_{i,j},u_{i,j}) = (x_g - x_{i,j})^T (x_g - x_{i,j}) + \frac{1}{2} u_{i,j}^T R u_{i,j}$, where $R$ is a positive definite control penalty matrix.
We define the weight of $i^{th}$ trajectory $\omega_i$ as:
\begin{equation*}
    \omega_i = \exp(-\frac{1}{\lambda}(S_i)),
\end{equation*}
where $\lambda$ is the parameter that decides how much we trust the better-performed trajectories. Then the MPPI algorithm updates the control input using the following equation
\begin{equation}
    u_j = \frac{\sum_{i=1}^K \omega_{i} u_{i,j} }{ \sum_{i=1}^K \omega_{i} }
\end{equation}
for $j = t,\cdots,t+T-1$, which approximates the optimal control inputs using sampled trajectories.

In conclusion, the MPPI algorithm uses sample trajectories to find the optimal control input to solve the motion planning problem. Because the MPPI algorithm avoids calculating the derivatives of the nonlinear dynamic systems or the value functions, it can be implemented in real-time with the help of parallel computations on the GPUs, even for complex dynamic systems.

\begin{figure}
    \centering
    \includegraphics[width=0.45\textwidth]{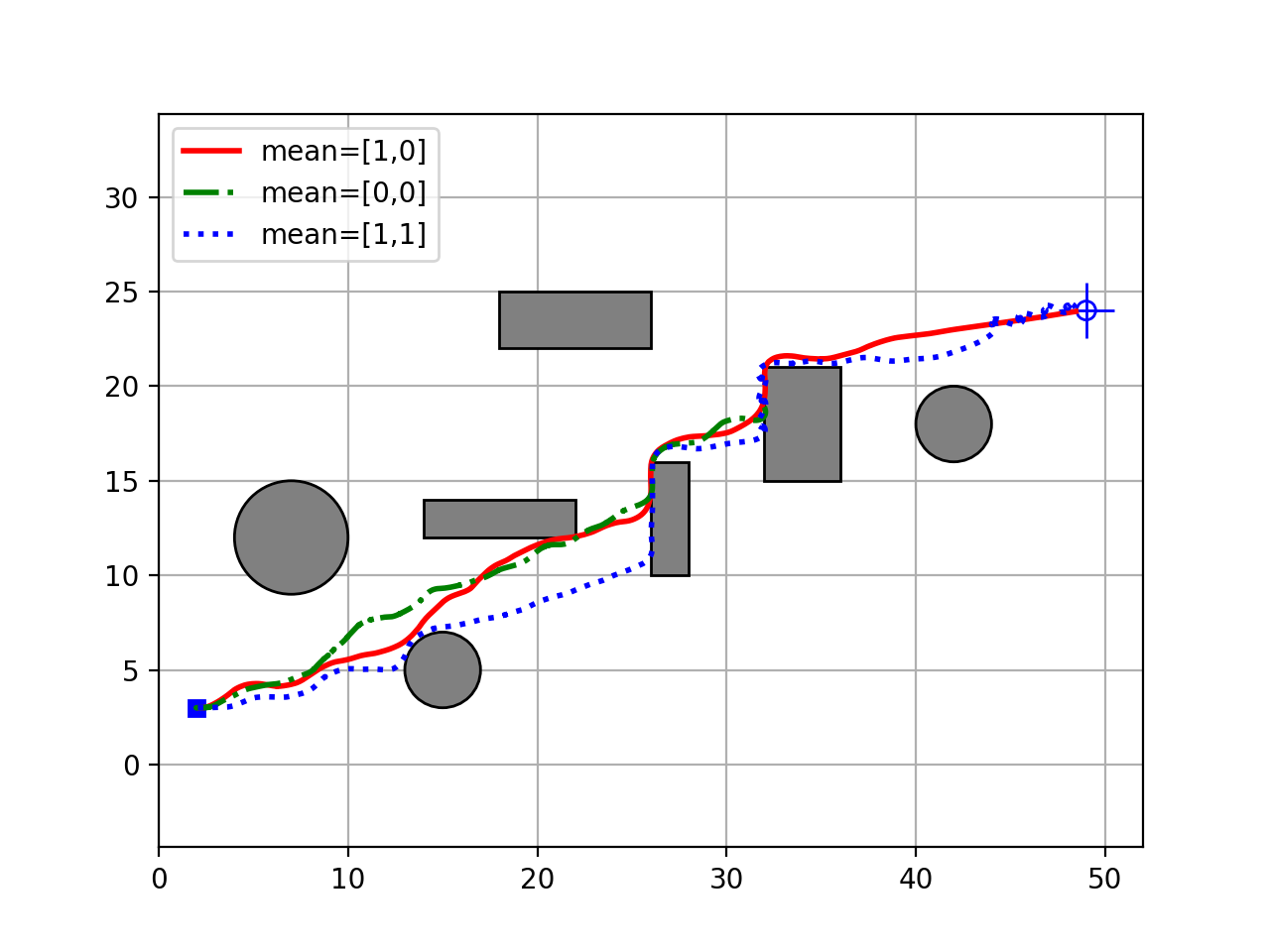}
    \caption{MPPI algorithm with various mean values in a static environment.}
    \vspace{-0.5cm}
    \label{fig:mppi1}
\end{figure}

While the MPPI algorithm has clear merits mentioned in the previous paragraph, its performance is dramatically influenced by the mean of the control input distribution. In the unicycle simulations presented in figure \ref{fig:mppi1}, the MPPI algorithm may fail to solve the motion planning problems due to the bad choice of the mean value. In figure \ref{fig:mppi1}, the red path is the result when the mean value of the control input is $\mu =[1,0]^T$, the yellow path is the result when the mean value is $\mu =[1,1]^T$, and the green path is the result when the mean value is $\mu =[0,0]^T$. We can find out easily that a smaller mean value hinders the exploration and cannot finish the path planning task in the provided horizon. A larger mean value may result in a safety violation. If the mean value is large, the MPPI algorithm is more aggressive and finishes the task faster. However, it also provides more risky control inputs and should require a larger sample size to get an optimal solution. We will show a required sample size analysis in Section~\ref{sec:analysis}.

\begin{figure*}[ht]
     \centering
     \begin{subfigure}[b]{0.48\textwidth}
         \centering
         \includegraphics[width=0.9\textwidth]{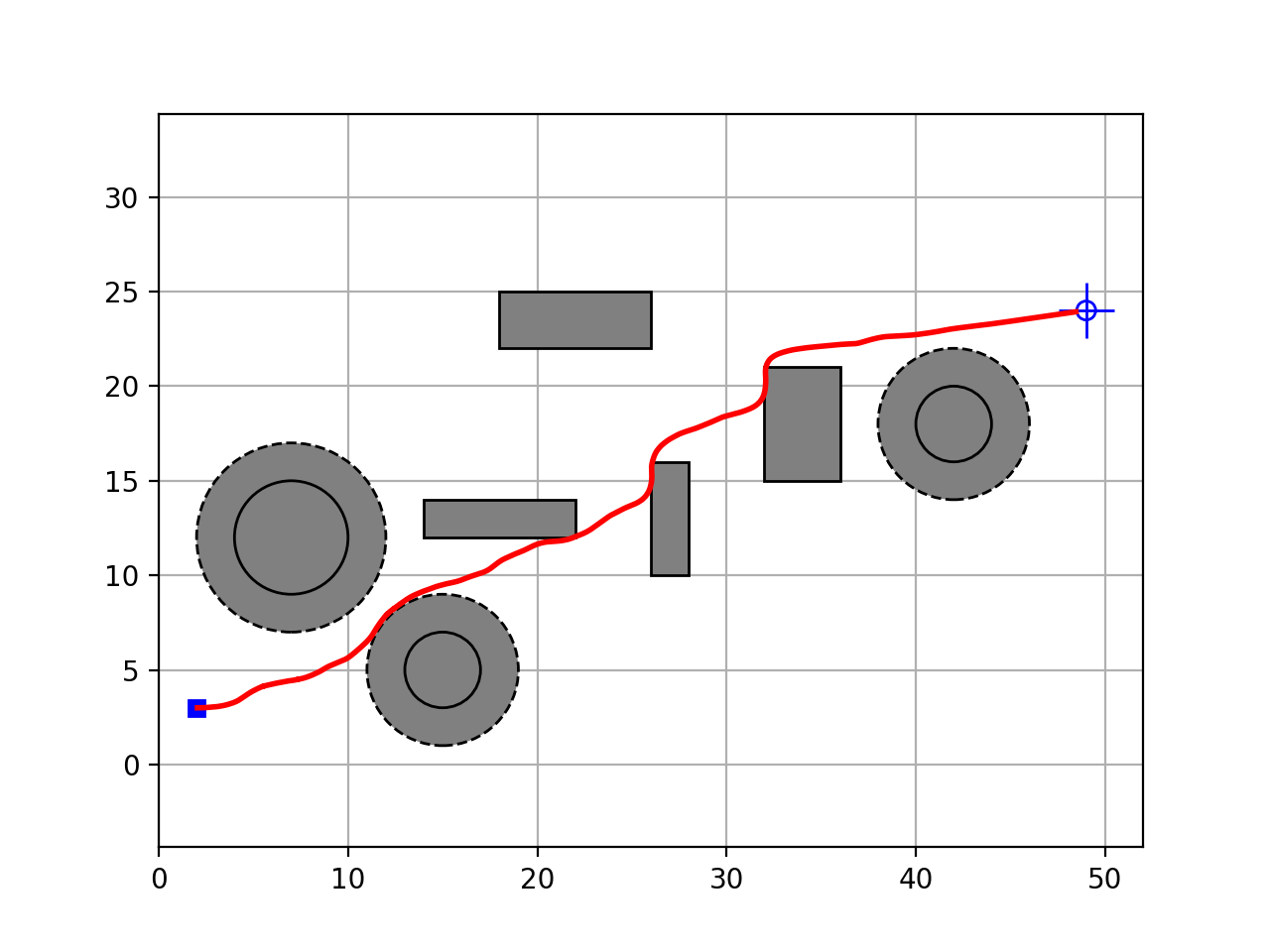}
     \end{subfigure}
     \hfill
     \begin{subfigure}[b]{0.48\textwidth}
         \centering
         \includegraphics[width=0.9\textwidth]{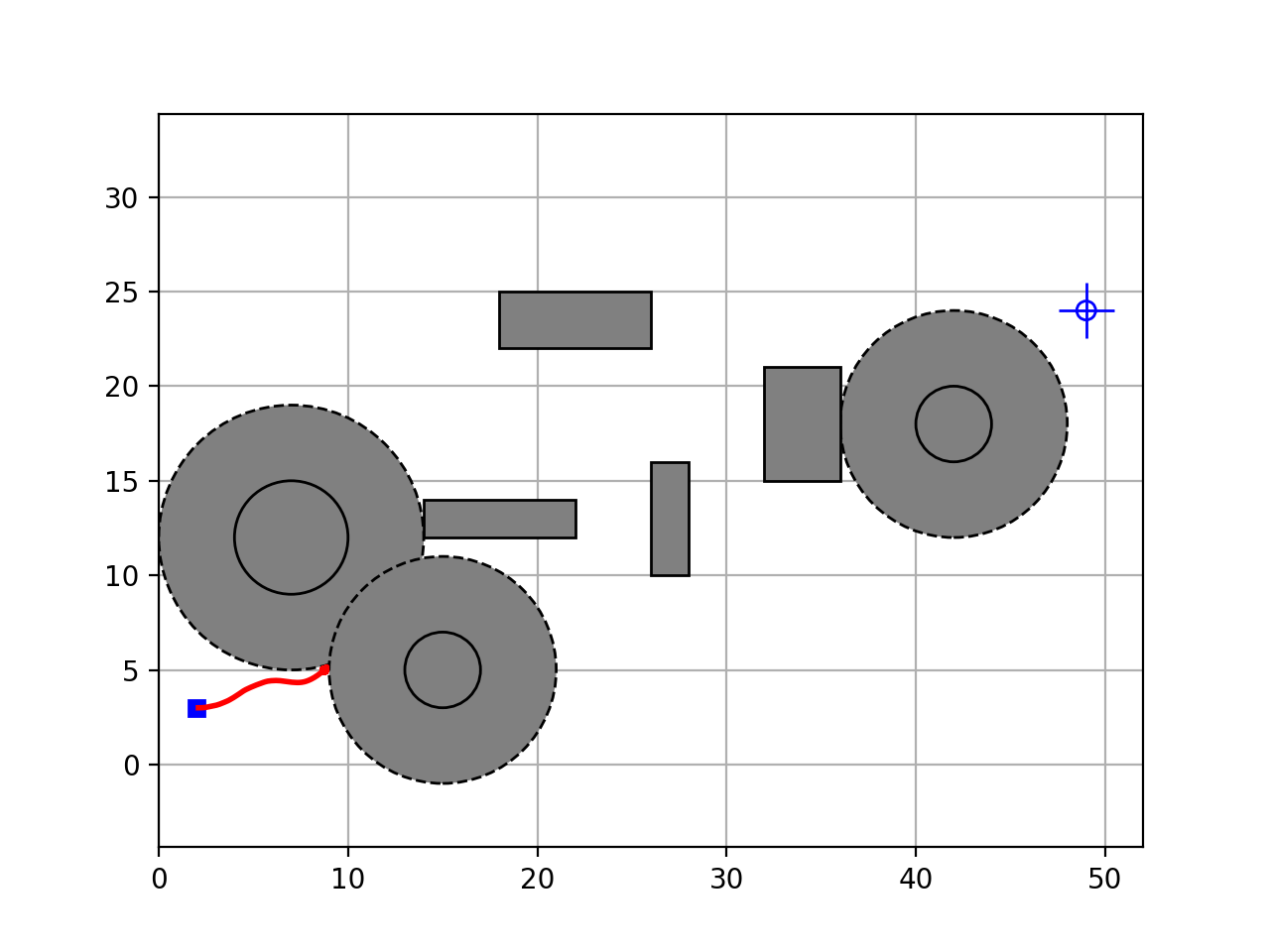}
     \end{subfigure}
        \caption{Results with MPPI algorithm in dynamic environments. In the first environment, the radius of the circle obstacles increases by 2 at time $t = 0.5s$. The solid line represents the radius before the environment changes, and the dotted line represents the radius after the environment changes. The MPPI algorithm can still handle the change without changing the mean value. In the second environment, the radius increases by 4 at time $t=0.5s$ as well. The MPPI algorithm with a fixed mean value fails to finish the task.}
        \vspace{-0.6cm}
        \label{fig:mppi2}
\end{figure*}

An example of MPPI trajectory in a dynamic environment is shown in Figure \ref{fig:mppi2}, where we increase the radius of circle obstacles by $2$ and $4$. The mean value for the MPPI is $[1,0]^T$, which has a perfect performance in a static environment but fails the task, being stuck in between obstacles in dynamic environments, as the second figure shown in Figure \ref{fig:mppi2}. Thus, we can conclude that for a dynamic environment, the MPPI algorithm needs a time-varying mean value to obtain a fine performance. To automate the procedure of choosing dynamic mean values, we combine it with the RRT algorithm.

\subsection{Replanning RRT Guided MPPI Algorithm} \label{sub:RRT-MPPI}
We propose a new sampling-based method that utilizes the RRT algorithm to guide the MPPI algorithm to solve the optimal motion planning problem defined in Problem \ref{pro1}. Our algorithm performs well without tuning the mean value of the control distribution. Our algorithm also has a fast running speed, and thus it can be implemented in real-time. 

First, we use the RRT algorithm to provide an offline nominal path $p_n$, which provides a possible solution to solve the motion planning problem. Although the RRT algorithm has a relatively fast iteration speed, the algorithm is still hard to  implement in real-time, which will also be shown later in the simulations. So we first run the RRT algorithm offline, not in real-time. Since the RRT algorithm only provides the state information instead of the control information, we then use Lyapunov controllers or PD controllers to obtain a nominal control input $u_n$ in real-time. However, since the RRT algorithm cannot guarantee the optimal solution, we use the MPPI algorithm with nominal control input $u_n$ as the mean of the random control distribution to explore an optimal control input $u^*$ at each time step. 

As the difference between the nominal control input $u_n$ and the optimal control input $u^*$ becomes increasingly large, the optimal control input may lead the agents to reach the area where the path $p_n$ of the RRT algorithm never reached before. In this case, the nominal path may have a negative influence on the MPPI algorithm. To address this issue, we use the replanning idea first presented in the DRRT algorithm, where the agents replan under the changing environment. However, unlike the DRRT algorithm, our algorithm replans when the nominal controllers $u_n$ are no longer helpful. In our implementation, we use a distance $R$ to justify if the replanning is needed. We use a \textit{NearestNeighbor}() function to calculate the distance between the current state $x$ and its closest point $x_n$ in the nominal path, and if the distance is larger than $R$, our algorithm replans. 

The detail of the Replanning RRT algorithm is presented in algorithm \ref{Replan_alg}. Our replanning RRT algorithm has a different input compared to the previous RRT algorithm, where the vertices set $\mathcal{S}$ is given by the previous nominal path $p_n$, and vertices set $\mathcal{S}'$ contains the start state $s_s$. Next, we sample the state $s_{sample}$ and find the nearest neighbor $s_{near}'$ and project the states $s$  same way as the RRT algorithm  in Algorithm \ref{alg:RRT}. However, we will also find the closest state $s_{near}$ to the vertices set $\mathcal{S}'$. Then we check if the new edges $(s_{near}', s)$ are in the collision-free space $\mathcal{X}_f$. Finally, we repeat the previous procedure until the distance between new vertex $s$ and target state $s_g$ or closest state on the nominal path $s_{near}$ is smaller than the radius $\gamma$ and return the new path $p$. Since our replanning algorithm uses the MPPI algorithm to give a penalty to the obstacles at each time step, we do not need to trim the previous result. As a result, the proposed algorithm is significantly faster than the original RRT algorithm and can be implemented on the robots in real-time.

\begin{algorithm}
\caption{Replanning RRT algorithm}\label{Replan_alg}
\begin{algorithmic}
\STATE{\textbf{Given}: Vertices $\mathcal{S} \leftarrow [p]$, $\mathcal{S'} \leftarrow [s_s]$}
\STATE{\textbf{Given}: Edges $\varepsilon \leftarrow \varnothing$}
\FOR{$i=1,...,n$}
    \STATE{$s_{sample} \leftarrow$\textit{SampleState}$()$;}
    \STATE{$s_{near}\leftarrow$\textit{NearestNeighbor}$(\mathcal{S}, s_{sample})$;}
    \STATE{$s_{near}'\leftarrow$\textit{NearestNeighbor}$(\mathcal{S}', s_{sample})$;}
    \STATE{$s\leftarrow$\textbf{Steer}$(s_{sample}, s_{near}',\gamma)$}
    \IF{\textit{ObstacleFree$(s_{near}', s)$}}
        \STATE{$\mathcal{S} \leftarrow \mathcal{S} \cup \left \{ s \right \}$;}
        \STATE{$\varepsilon \leftarrow \varepsilon \cup \left \{ (s_{near}, s) \right \}$;}
    \ENDIF
    \IF{$d(s,s_g)<\gamma$ or $d(s, s_{near})<\gamma$}
        \STATE{$\mathcal{S}' \leftarrow \mathcal{S}' \cup \left \{ s_g \right \}$ or $\mathcal{S}' \leftarrow \mathcal{S}' \cup \left \{ s_{near} \right \}$;}
        \STATE{$\varepsilon \leftarrow \varepsilon \cup \left \{ (s, s_g) \right \}$ or $\varepsilon \leftarrow \varepsilon \cup \left \{ (s, s_{near}) \right \}$;}
        \STATE{\textbf{return}  $p=$ \textit{ExtractPath} $(\mathcal{S}', \varepsilon)$}
    \ENDIF
\ENDFOR    
\end{algorithmic}
\end{algorithm}

With the new nominal path $\mathcal{S}'$, we then use the Lyapunov controllers or the PD controllers to get the new nominal control input $u_n'$. We can obtain a sampled trajectory $\tau_{i} = [ x_{i,t}, ..., x_{i,t+T-1}]^T$ with the new distribution $\mathcal{N}(u_n', \Sigma)$, where $T$ is the time horizon of the MPPI algorithm and $\Sigma$ is the fixed variance. Then we calculate the cost of the $i^{th}$ sampled trajectory by using the quadratic cost function $ S(\cdot)$ and using the following equation, we calculate the weight of each trajectory:
\begin{equation}\label{eq:sample_weight}
    \omega_{i} = \exp \left (-\frac{1}{\lambda} S(\tau_{i}) - \min(S(\tau_i)) \right ).
\end{equation}

Note that we need to find the minimum value of all trajectories to prevent the numerical instability of the algorithms~\cite{yin2022trajectory}. Finally, we use the normalized trajectory weights to calculate the control update law: for $j=t,\cdots,t+T-1$,
\begin{equation} \label{eq:control_update}
    u_j =\frac{\sum_{i=1}^K \omega_{i} u_{i,j} }{ \sum_{i=1}^K \omega_{i} }.
\end{equation}

The proposed RRT-guided MPPI algorithm is summarized in Algorithm~\ref{algo1}.
\begin{algorithm}
\caption{RRT-MPPI algorithm}
\begin{algorithmic}\label{algo1}
\STATE{\textbf{Given:} Number of sample trajectories $K$ and timesteps $T$;}
\STATE{\textbf{Given:} Initial variance $\Sigma_0$;}
\STATE{\textbf{Given:} Cost function parameters $\phi, q, R, \lambda$;}
\WHILE{task is not completed}
    \STATE{Use RRT algorithm to get initial path $p_n$}
    \FOR{$j \leftarrow t$ to $t+T-1$}
        \STATE{Find the nearest state $s\in p_n$ to the current state $x$;}
        \IF{$d(s,x)\geq R$}
            \STATE{Use Replanning RRT algorithm to get new $p_n$}
            \STATE{Find new nearest state $s \in p_n$}
        \ENDIF
        \STATE{Get nominal control mean value $u_n = L(s,x)$}
        \FOR{$i \gets 0$ to $K-1$}
            \STATE{Generate control variations $u_{i,j}$ $\sim \mathscr{N}(u_n,\Sigma_0)$;}
            \STATE{Simulate discrete dynamic \eqref{eq:discrete_dynamic} to obtain $x_{i,j}$;}
            \STATE{Calculate cost function $S(\tau_{i}) \mathrel{+}= q(x_{i,j}, u_{i,j})$;}
        \ENDFOR
        \STATE{Calculate the terminal cost $S(\tau_i) \mathrel{+}= \phi( {x_{i,t+T}})$}
    \ENDFOR
    \STATE{$\beta \gets \min_i[S(\tau_{i})]$;}
    \STATE{Get sample weights $\omega_{i}$;}
    \STATE{Update control input using $\omega_{i,j}$ and $u_{i,j}$;}
    \STATE{Send $u_{t}$ to actuator;} 
\ENDWHILE
\end{algorithmic}
\end{algorithm}

\vspace{-0.4cm}
\section{Sample Size Analysis}\label{sec:analysis}

In this section, we show that if the mean value of the MPPI increases, the required sample size also increases. On the other hand, a sufficiently large sample size is required to explore the free space, as shown in Section~\ref{sub:MPPI}. As a result, we emphasize the necessity of an automated design procedure for a time-varying mean value by the RRT algorithm.

In the MPPI algorithm, the sample size significantly influences the performance and running time. We provide the analysis of the required sample size of the MPPI algorithm based on the error between optimal control input provided by the Hamilton-Jacobi-Bellman (HJB) equation and its sampling-based approximation.

The MPPI algorithm~\cite{williams2018information} solves an optimal control problem with a quadratic control cost and a state-dependent cost. The corresponding value function $V(x_t)$ is then defined as:
\begin{equation}\label{eq:stochastic_prob}
    \min_{u_t} \mathbb{E}\left[ \phi(x_T) + \int_t^{T\Delta t}q'(x_t,u_t)dt \right],
\end{equation}
where $q'$ is the continuous-time version of the cost-to-go in~\eqref{eq:prob} and $\Delta t$ is the sampling time.
With the boundary condition $V(x_T) = \phi(x_T)$, the solution to the stochastic HJB equation~\cite{stengel1994optimal, fleming2006controlled} for the system defined in \eqref{eq:dynamics} and the value function defined in \eqref{eq:stochastic_prob} is given as follows:
\begin{equation}
    u^*(x_t) = -R_t^{-1} g(x_t)^T V(x_t).
\end{equation}
However, the Partial Differential Equations (PDEs) are hard to solve for the curse of dimensionality. Thus, the algorithm uses exponential transformation of the value function $V(x_t) = -\lambda \log(\Phi(x_t))$ to make the HJB linear in $\Phi$. The linearity allows us to solve the problem with forward-sampling. The iterative control update law can be calculated as a ratio of the expectations (the detailed derivations, see~\cite{kappen2005path, theodorou2010generalized}):
\begin{equation}\label{eq:optimal}
    u(x_t)^*=\frac{\mathbb{E}[\exp (-(1/\lambda)) S(\tau)\sigma(x_t) d W_t ]}{\mathbb{E}[\exp (-(1/\lambda) S(\tau) ]},
\end{equation}
where $S(\tau) = \phi(x_T)+ \int_{t_0}^{T\Delta t} q'(x_t,t)dt$, and $\tau$ is a random trajectory process. The continuous-time trajectories are sampled as a \emph{discretized} system according to
\begin{equation}\label{eq:discrete_dynamic}
    dx_t = \left(f(x_t) + g(x_t)u_t \right) \Delta t + \sigma(x_t) \delta_t \sqrt{\Delta t},
\end{equation}
where $\delta_t$ is the time-varying vector of standard normal Gaussian random variables, and $\Delta t$ denotes the time step of the time-discretization, and we use Euler–Maruyama method~\cite{platen2010numerical}. Then the discrete-time control update law to approximate the optimal control becomes
\begin{equation}\label{eq:discrete_control}
    u_j\approx  \frac{\sum_{i=0}^{K-1} \exp(-(1/\lambda) S (\tau_{i}))\delta u_{i,j} }{\sum_{i=0}^{K-1} \exp(-(1/\lambda) S (\tau_{i}))},
\end{equation}
where $\delta u_{i,j}$ can be considered as a random control input, and $S(\tau)=\phi(x_{i,t+T})+\sum_{j=t}^{t+T-1} q(x_{i,j},\delta u_{i,j})$.

In conclusion, the MPPI algorithm uses Monte Carlo (MC) methods to approximate the optimal control solution \eqref{eq:optimal} with the sampling-based control input \eqref{eq:discrete_control}. Our previous works on the sampling complexity of the MPPI method~\cite{yoon2022sample, tao2022path} use Hoeffding's inequality and Chebyshev's inequality to provide the required sample size given error bounds and risk probability. Compared to the previous work, we focus on the influence of the mean of the sampling control distribution instead of the variance and prove that a larger mean of the random control distribution requires a larger sample size. Note that, while we only discuss the case of one-dimensional control input $\delta u_j \sim \mathcal{N}(\mu, \Sigma)$ for notational simplicity, the result can be extended to high-dimensional control input straightforwardly.

We divide the expectation of the control input \eqref{eq:discrete_control} to two different parts and the define each part to be $E_1$ and $E_2$, the definitions of $E_1$ and $E_2$ are expressed as:
\begin{equation}
    \begin{aligned}
        E_1 &= \E[\omega] = \E [\exp (-\frac{1}{\lambda} S(\tau_i))], \\
        E_2 &= \mathbb{E} \left [ \frac{\omega \delta u}{\mathbb{E}[\omega]} \right ].
    \end{aligned}
\end{equation}

The MC integration used in the MPPI algorithm to approximate those terms are defined as $\hat E_1$ and $\hat E_2$:
\begin{equation}
    \begin{aligned}
        \hat E_1 &= \sum_{i=1}^T \exp(-\frac{1}{\lambda} S(\tau_i)), \\
        \hat E_2 &= \frac{1}{K} \sum^K_{i=1} (\frac{\omega_i \delta u_i}{\E [\omega]}).
    \end{aligned}
\end{equation}

We define the error bounds $\epsilon_1, \epsilon_2$ and the risk probability $\rho_1, \rho_2$ by:
\begin{equation}
    \begin{aligned}
        &\mathbb{P}\left \{ \left | \hat E_1 - E_1 \right | \geq \epsilon_1 \right \} \leq \rho_1, \\
        &\mathbb{P}\left \{ \left |\hat E_2-E_2\right |\geq \epsilon_2 \right \}  \leq \rho_2 .
    \end{aligned}
\end{equation}

We make the following assumptions for the MPPI algorithm:
\begin{ass}\label{ass:cost}
We suppose that the running cost function $q(x_{i,t},u_{i,t})$ and terminal cost function $\phi(x_{i,t+T})$ are quadratic functions.
\end{ass}

\begin{ass}\label{ass:1}
Assume that the error bound $\epsilon_1$ of Chebyshev's inequality is smaller than the expectation of $\omega$
\vspace{-0.1cm}
\begin{equation*}
    \epsilon_1 < \E [\exp (-\frac{1}{\lambda} S(\tau_i))].
\end{equation*}
\end{ass}

\begin{The}\label{thm:thm1}
If Assumptions \ref{ass:cost} and \ref{ass:1} are satisfied, and given the same sampling complexity error bounds $\epsilon_1, \epsilon_2$ and risk probabilities $\rho_1, \rho_2$, the required sample size $K=\max(K_1, K_2)$ becomes larger as the mean of random control distribution $\E[\delta u]$ becomes larger.
\end{The}

To prove the main theorem, we need the following intermediate results, lemmas.
\begin{lemma}\label{lemma:1}
(\cite{tao2022path}) For any random variables $X, Y$, we have:
\vspace{-0.1cm}
\begin{equation*}
    \Var \left [ XY \right]   \leq 2\Var[X] \Var[Y] + 2 \Var [Y] \E[X]^2 .
\end{equation*}
\end{lemma}

\begin{lemma}
It holds that
$\Var [\omega] \leq (1-\E [\omega]) \E [\omega] \leq \E [\omega] \leq 1$.
\end{lemma}
\vspace{0.2cm}
\begin{proof}
Since $\omega = \exp(-\frac{S(\tau)}{\lambda})$ and since the cost-to-go function $S(\tau) \geq 0$ by Assumption \ref{ass:cost}, then $\omega \in [0,1]$ is a bounded random variable and its variance is also bounded.
\end{proof}

With the above lemmas, we have the following proof for Theorem \ref{thm:thm1}:

\begin{proof}
We first find the required sample size $K_1$. Since $\omega \in [0,1]$, using Hoeffding's inequality, we have:
\begin{align*}
   \mathbb{P} \{ | \hat E_1 - \E [\omega] | \geq \epsilon_1 \} &\leq 2 \exp \left ( -\frac{K_1\epsilon_1^2}{(\omega_{\max}-\omega_{\min})^2} \right )
   \\&\leq 2\exp(-K_1\epsilon_1^2) .
\end{align*}
By the definition of risk probability $\rho_1$, we have:
\begin{equation}\label{eq:N1}
    \mathbb{P} \{ | \hat E_1 - \E [\omega] | \geq \epsilon_1 \} \leq \rho_1 = 2\exp(-K_1\epsilon_1^2),
\end{equation}
So the sample size $K_1$ has the following formulation:
\begin{equation}\label{eq:K_1}
    K_1 = -\frac{1}{\epsilon_1^2}  \log{\frac{\rho_1}{2}}.
\end{equation}
Next, we calculate the required sample size $K_2$. Using Chebyshev's inequality, we have:
\begin{align*}
        &\mathbb{P} \left \{ \left | E_2 - \hat E_2 \right | \geq \epsilon \right \} \leq \rho_2 = \frac{\text{Var}[E_2]}{K_2 \epsilon_2^2} .
\end{align*}
Next, we find an upper bound of the variance of $E_2$:
\begin{align*}
    \Var [E_2] &= \Var \left [ \frac{\omega  \delta u_t}{\E [\omega]} \right ]\\
    &= \frac{1}{\E [\omega]^2} \Var \left [ \omega[\delta u_t] \right ] \\
    &\leq \frac{2 \Var[\omega] \Var[\delta u] + 2 \Var [\omega] \E[\delta u]^2 }{\E [\omega]^2} \\
    &\leq \frac{2 (\Var [\delta u] + \E[\delta u]^2)}{\E [\omega]^2} ,
\end{align*}
where the first inequality uses the result of Lemma \ref{lemma:1},  and since $\omega = \exp (-\frac{1}{\lambda} S(\tau) )$, then we can obtain that $\omega \in (0,1)$. So we can conclude that $var[\omega] \leq 1$. Thus, Chebyshev's inequality leads to:
\begin{equation} \label{eq:N2}
    \mathbb{P} \left \{ \left | E_2 - \hat E_2 \right | \geq \epsilon \right \}\leq \rho_2 = \frac{\Gamma}{K_2\epsilon_2^2 \E [\omega]^2},
\end{equation}
where $\Gamma = 2 (\Var [\delta u] + \E[\delta u]^2)$. The expectation of the sample weights $\E[\omega]$ is not easy to be calculated, but it can be derived from the previous result where $| \hat E_1 - \E [\omega] | \geq \epsilon_1$, so we have the following result:
\begin{equation*}
    \frac{1}{(\hat E_1+\epsilon_1)^2} \leq \frac{1}{(\E[\omega])^2} \leq \frac{1}{(\hat E_1-\epsilon_1)^2}.
\end{equation*}
So the required sample size $K_2$ is:
\begin{equation}\label{eq:K_2}
    K_2 = \frac{\Gamma}{\rho_2\epsilon_2^2}\left (\frac{1}{\hat E_1 - \epsilon_1}\right )^2.
\end{equation}
From equation \eqref{eq:K_1}, we can conclude that the required sample size $K_1$ is the same when the error bounds and risk probability are the same. Thus, the term $\hat E_1$ is independent of the choice of the mean of the control distribution $\E[\delta u]$. From equation \eqref{eq:K_2}, we can find out that term $\Gamma =  2 (\Var [\delta u] + \E[\delta u]^2)$ decides the number of the required samples. A larger mean $\E [\delta u]$ means larger $\Gamma$ value.
\end{proof}

The main Theorem \ref{thm:thm1} in this section shows the necessity of designing a time-varying mean value for the random control input distribution. From the previous discussion, we can conclude that a larger mean value results in more aggressive exploration behavior, and we need to sample more trajectories to approximate the optimal control input. On the contrary, a smaller mean value results in conservative exploration and cannot finish the task in the scheduled time. In a dynamic system, the mean value needs to be tuned as the environments change. Thus, we use an RRT-based mean value to guide the MPPI algorithm to provide a better performance in solving the optimal motion planning problems in dynamic environments.

\section{Simulations}\label{sec:simulations}
\subsection{Unicycle Dynamics}
We implement our algorithm on a two-dimensional unicycle dynamic system with:
\begin{equation*}
    \begin{bmatrix}
    \dot x^d \\ 
    \dot y^d \\ 
    \dot \theta\\
    \dot \phi
    \end{bmatrix} 
    = \begin{bmatrix}
    \cos \theta & 0\\ 
    \sin \theta & 0\\
    \frac{\tan \phi}{L} & 0 \\
    0 & 1
    \end{bmatrix}\begin{bmatrix}
    v + \delta^v \\ 
    \omega + \delta^\omega
    \end{bmatrix},
\end{equation*}
where $x, y$ are the coordinates, $\theta$ is the heading angle, and $\phi$ is the steering angle. $v$ is the linear velocity control input, and $\omega$ is the angular velocity control input. $L=0.5$ is the length of the wheelbase. $\delta = [\delta^v, \delta^\omega]\sim \mathcal{N}(\bar 0,I)$ is the random control input perturbation. The time step for the discrete-time simulation is $\Delta t = 0.05 s$. We use the following discrete dynamics in the MPPI algorithm:
\begin{equation*}
    \begin{bmatrix}
    x^d_{t+1} \\ 
    y^d_{t+1} \\ 
    \theta_{t+1}\\
    \phi_{t+1}
    \end{bmatrix} 
    =
    \begin{bmatrix}
    x^d_{t} \\ 
    y^d_{t} \\ 
    \theta_{t}\\
    \phi_{t}
    \end{bmatrix} + \Delta t \begin{bmatrix}
    \cos \theta_t & 0\\ 
    \sin \theta_t & 0\\
    \frac{\tan \phi_t}{L} & 0 \\
    0 & 1
    \end{bmatrix}\begin{bmatrix}
    v_t + \delta_t^v \\ 
    \omega_t + \delta_t^\omega
    \end{bmatrix},
\end{equation*}

\subsection{Simulation Setups}
The maximum sample size in the RRT algorithm is set to $20000$, and the projection radius is set to $\gamma=0.5$.
We set the sample size for the MPPI algorithm to be $K=10000$, the time horizon to be 20, and $\lambda = 1.0$. The cost function is defined as:
\begin{equation*}
    q(x) = \| x- x_g\|_2^2 + 1000 * \mathds{1}_{x\in \mathcal{X}_{obs}},
\end{equation*}
where $x$ represents the current states, and $x_g$ represents the goal state. $\mathcal{X}_{obs}$ is the obstacle set over $\mathbb{R}^2$, and $\mathds{1}$ is the indicator function. We test our algorithm in two different environments, a static environment, and a dynamic environment. In both simulations, the start states are $x_s = [2,3,0,0]^T$, and the goal states are $x_g = [49,24,0,0]^T$. We use a Lyapunov controller to design the velocity control input and a Proportional controller to design the angular velocity control input:
\begin{equation}
\begin{aligned}
    u_v &= e_d v_{max}  \frac{(1 - \exp (-\alpha \| e_d \|^2 ))}{\|e_d\|}, \\
    u_\omega &= k_p  e_{\theta},
\end{aligned}
\end{equation}
where $e_d, e_\theta$ are the error between the desired target state and current states.

\subsection{Results}
\begin{figure*}[ht]
     \centering
     \begin{subfigure}[b]{0.32\textwidth}
         \centering
         \includegraphics[width=1.0\textwidth]{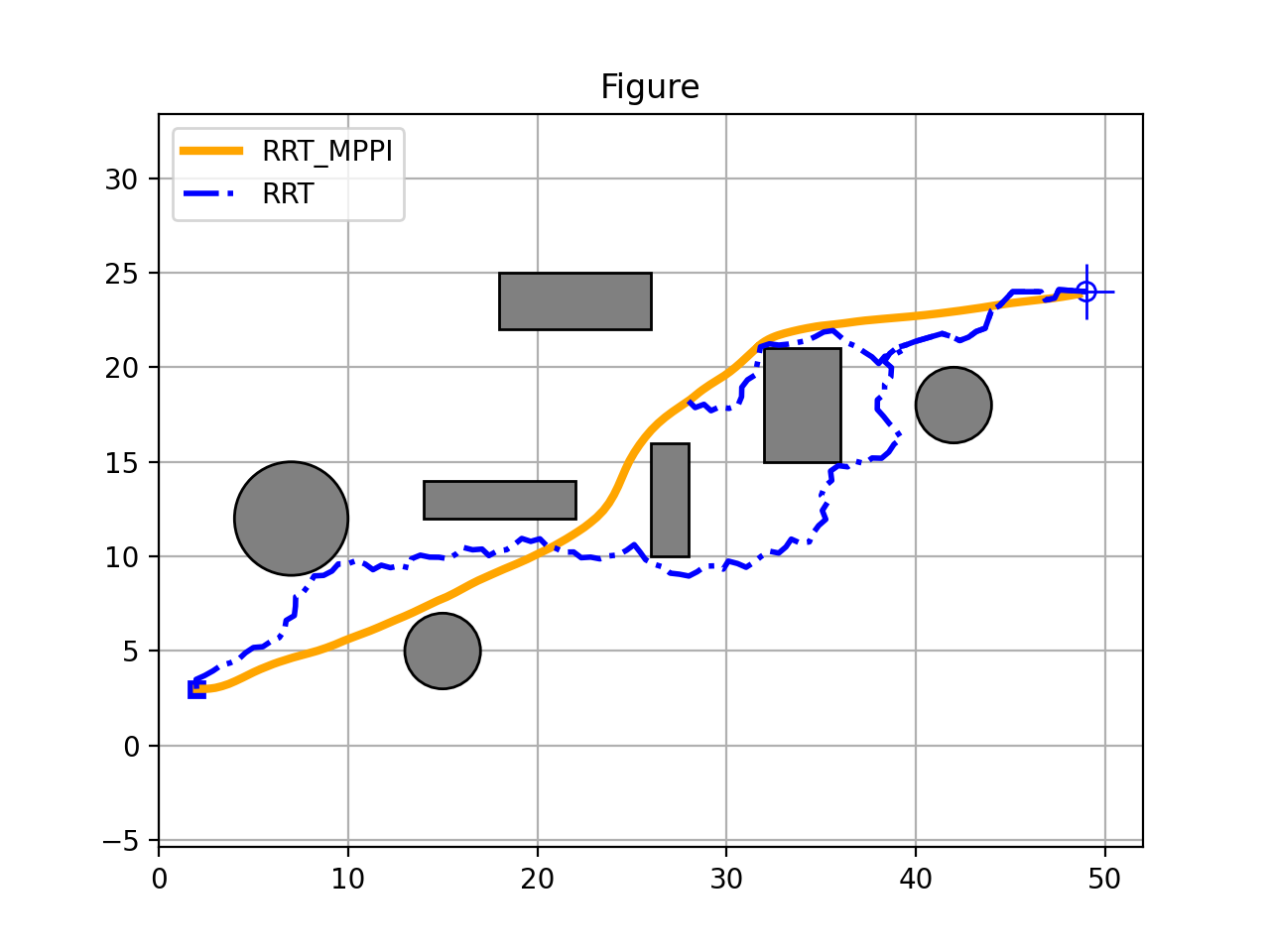}
         \caption{Static Environment}
         \label{fig:rrt_mppi1}
     \end{subfigure}
     \hfill
     \begin{subfigure}[b]{0.32\textwidth}
         \centering
         \includegraphics[width=1.0\textwidth]{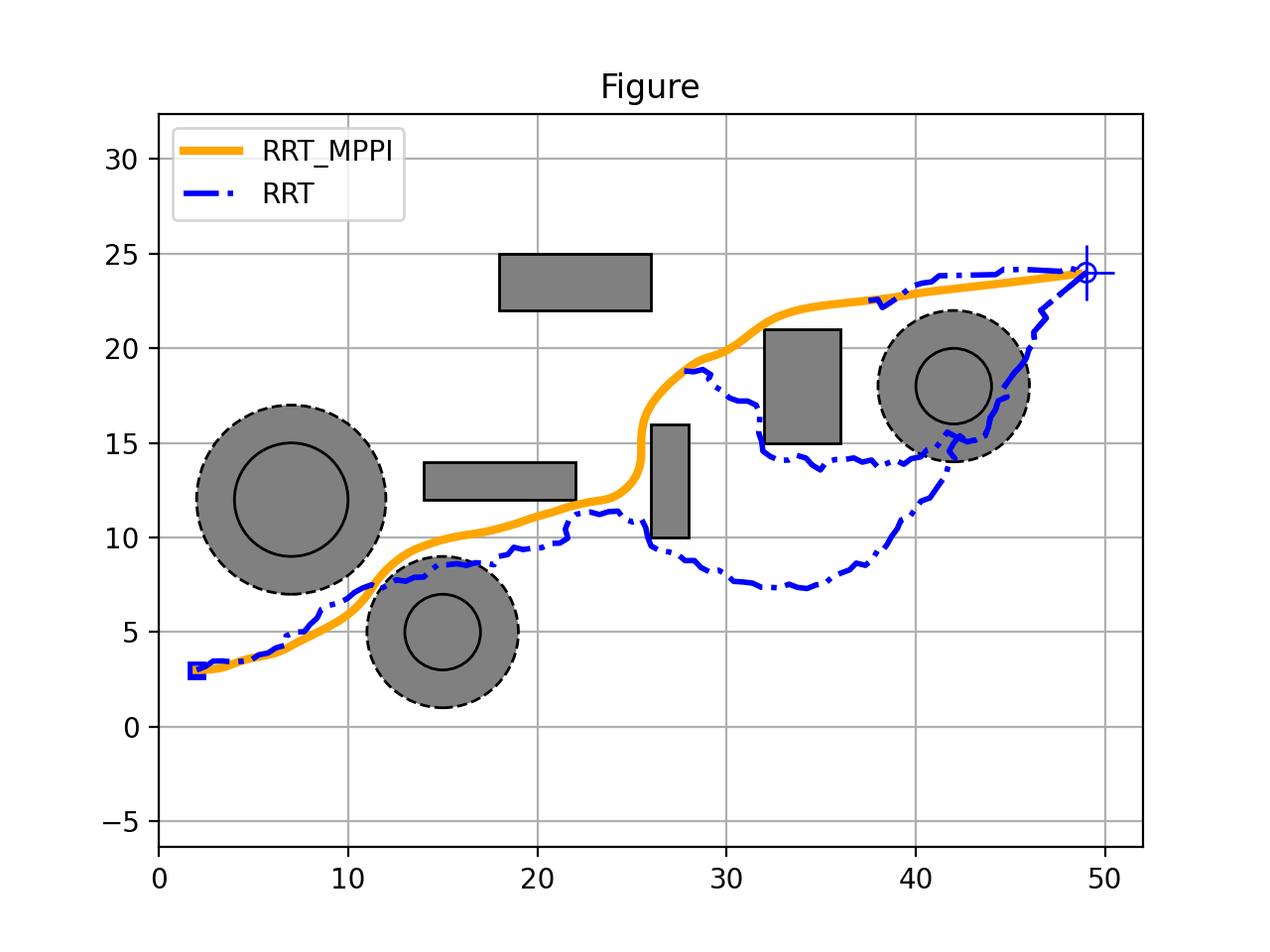}
         \caption{1st Dynamic Environment}
         \label{fig:rrt_mppi2}
     \end{subfigure}
     \hfill
     \begin{subfigure}[b]{0.32\textwidth}
         \centering
         \includegraphics[width=1.0\textwidth]{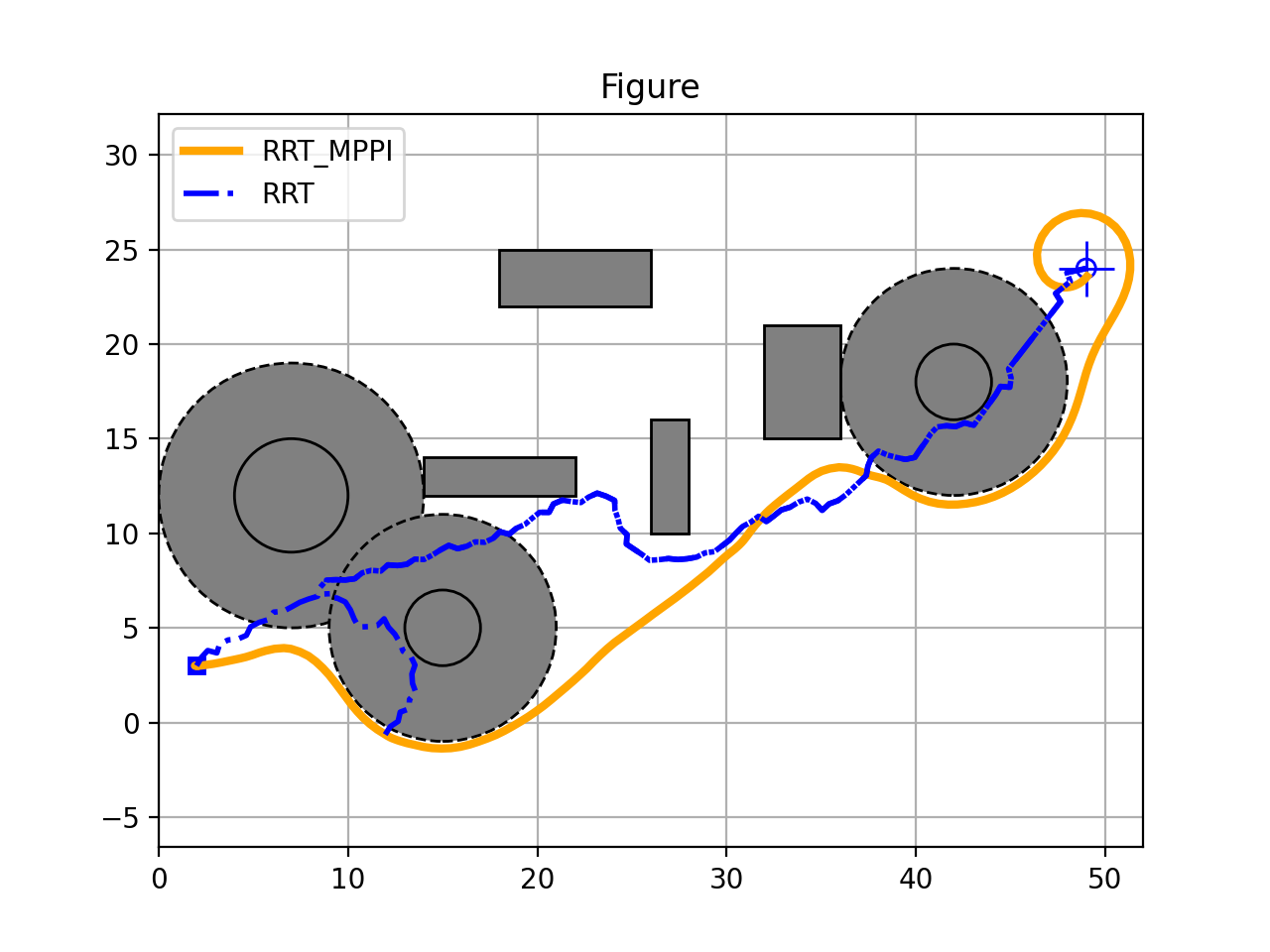}
         \caption{2nd Dynamic Environment}
         \label{fig:rrt_mppi3}
     \end{subfigure}
        \caption{Results with RRT-MPPI algorithm in static or dynamic environments. The blue dash-dot lines are the path provided by the RRT algorithm, and the orange line is the result of our proposed method.}\vspace{-0.7cm}
        \label{fig:2nd_exp}
\end{figure*}

We first test our algorithm in a fully known static environment with the replanning conditions $R=6$. Figure \ref{fig:rrt_mppi1} shows the result of a unicycle robot navigating through the obstacles. The black rectangles represent the boundary of the environments, the grey circles and rectangles denote the obstacles, the blue square denotes the start state $x_s$, and the blue cross denotes the goal state $x_g$. The blue line in the figures is the result of the replanning RRT path, and the orange line in the figures is the resulting control output from the RRT-MPPI algorithm. Next, we implement our algorithm in dynamic environments where the radius of the circle obstacles increases by $2$ and $4$. We plot  the environment changes by plotting the circles with dot lines as their boundaries. Figure \ref{fig:rrt_mppi2} and Figure \ref{fig:rrt_mppi3} show that our algorithm can handle dynamic environments. Note that in dynamic environments, the nominal path provided by the RRT path may violate safety. But, since the MPPI algorithm can explore freely, our algorithm is still able to find the solution to the optimal motion planning problems. Besides, we want to implement the algorithm in real-time, so the RRT algorithm we adopt here is relatively inaccurate and can only guide the MPPI algorithm.

\begin{figure}
    \centering
    \includegraphics[width=0.95\linewidth]{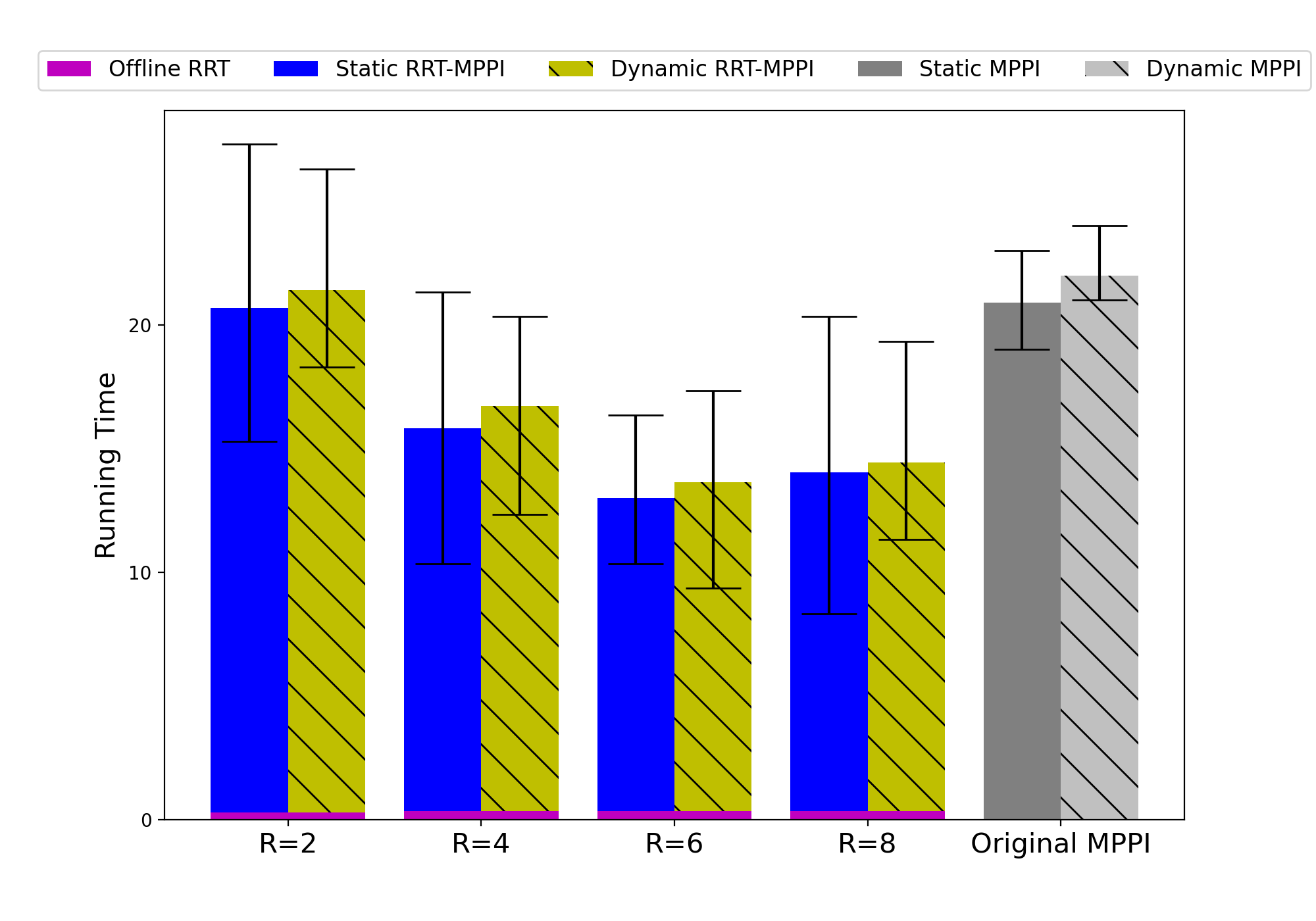}\vspace{-0.3cm}
    \caption{Running time of RRT-MPPI algorithm with different replanning conditions and MPPI algorithms with the fixed mean value. }
    \vspace{-0.7cm}
    \label{fig:bar}
\end{figure}

We repeat the previous experiments for 10 times and change the value of the replanning condition from $R=2$ to $R=8$. In  Figure \ref{fig:bar}, we plot  the average time, the maximum and minimum running time of our algorithm, and the original MPPI algorithm with mean $[1,0]^T$ in static and dynamic environments. The time of the offline RRT algorithm is in purple color. Note that even the offline RRT algorithm takes around 0.2 seconds, but it is still not fast enough to be implemented in real-time. The online RRT-MPPI algorithm for the static environment is in blue color, and the dynamic environment is in yellow color. We also compare the computation time with the MPPI algorithm with a fixed mean value $[\mu_v, \mu_\omega]^T =[1,0]^T$, which is the grey color in the Figures. As the radius decreases, the RRT-MPPI algorithm can provide a more accurate nominal controller. But the times of the replanning procedure increase as well, and as a result, the total time to complete the task becomes longer. We can see that when the radius $R=6$, the algorithm takes the least time to finish the motion planning task. All experiments are done on a Macbook Air laptop with an M1 chip in real-time. 

To calculate the required sample size, we set the desired bound $\epsilon_1 = 0.02, \epsilon_2 =0.1$, and set the allowable risk of failure $\rho_1 = 0.05, \rho_2 = 0.1$. We calculate the numbers of samples $K_1$ and $K_2$ based on the equations \eqref{eq:K_1} and \eqref{eq:K_2} at time $T * \Delta t = 50 * 0.05s = 2.5s$. Table \ref{tab:table2} shows that the required sample size of our algorithm is smaller than the original MPPI algorithm with a fixed mean value of 1.

\begin{table}[ht]
    \centering
    \caption{
     Required Sample Size for RRT-MPPI and MPPI algorithm.}
    \begin{tabular}{c c c c c}
    \toprule
         Algorithm  & Time & $K_1$ &$K_2$ & running time \\ \midrule \midrule
          MPPI with fixed mean 1 & 2.5s & 9222 & 11413 & 21.81s \\ \midrule
        RRT-MPPI & 2.5s & 9222  & 6122 & 14.56s\\
         \midrule
    \end{tabular}
    \label{tab:table2}
\end{table}

\vspace{-0.7cm}
\section{Conclusion}\label{sec:conclusion}
This paper presents a real-time RRT-MPPI algorithm to solve the motion planning problem in different environments. The proposed algorithm advances the RRT algorithm in terms of dynamic environment navigation and optimality and reduces the need to fine-tune the mean value of the MPPI algorithm. In particular, we use the RRT algorithm to provide the suitable nominal control mean value for the random distribution in the MPPI algorithm, which helps us to avoid fine-tuning the mean value and balance the optimality and exploration. Finally, in the simulations, we use a unicycle robot to implement the algorithm in real-time in static and dynamic environments. We compare the running time and required sample size of our RRT-MPPI algorithm with the fixed value MPPI algorithm in the experiments, showing that our algorithm is faster and requires a smaller sample size.

\vspace{-0.2cm}

\bibliographystyle{ieeetr}
\bibliography{citation}

\end{document}